\documentclass[final]{cvpr}

\usepackage{times}
\usepackage{epsfig}
\usepackage{graphicx}
\usepackage{amsmath}
\usepackage{amssymb}
\usepackage[linesnumbered,ruled,vlined]{algorithm2e}
\usepackage{multirow}
\usepackage{verbatim}
\usepackage{algorithm2e}

\usepackage[pagebackref=true,breaklinks=true,colorlinks,bookmarks=false]{hyperref}

\def\cvprPaperID{****} 
\def\confYear{CVPR 2021}

\begin{document}

\title{Team I2R-VI-FF Technical Report on EPIC-KITCHENS VISOR Hand Object Segmentation Challenge 2023}

\author{
Fen Fang, Yi Cheng, Ying Sun and Qianli Xu\\
Institute for Infocomm Research, A*STAR, Singapore\\
{\tt\small \{fang\_fen, cheng\_yi, suny, Qxu\}@i2r.a-star.edu.sg}
}

\maketitle

\begin{abstract}
In this report, we present our approach to the EPIC-KITCHENS VISOR Hand Object Segmentation Challenge, which focuses on the estimation of the relation between the hands and the objects given a single frame as input. The EPIC-KITCHENS VISOR dataset provides pixel-wise annotations and serves as a benchmark for hand and active object segmentation in egocentric video. Our approach combines the baseline method, i.e., Point-based Rendering (PointRend) and the Segment Anything Model (SAM), aiming to enhance the accuracy of hand and object segmentation outcomes, while also minimizing instances of missed detection. We leverage accurate hand segmentation maps obtained from the baseline method to extract more precise hand and in-contact object segments. We utilize the class-agnostic segmentation provided by SAM and apply specific hand-crafted constraints to enhance the results. In cases where the baseline model misses the detection of hands or objects, we re-train an object detector on the training set to enhance the detection accuracy. The detected hand and in-contact object bounding boxes are then used as prompts to extract their respective segments from the output of SAM. By effectively combining the strengths of existing methods and applying our refinements, our submission achieved the 1st place in terms of evaluation criteria in the VISOR HOS Challenge.

\end{abstract}

\section{Introduction}

\begin{figure}[ht]
     \centering
     \includegraphics[width=1.0\linewidth]{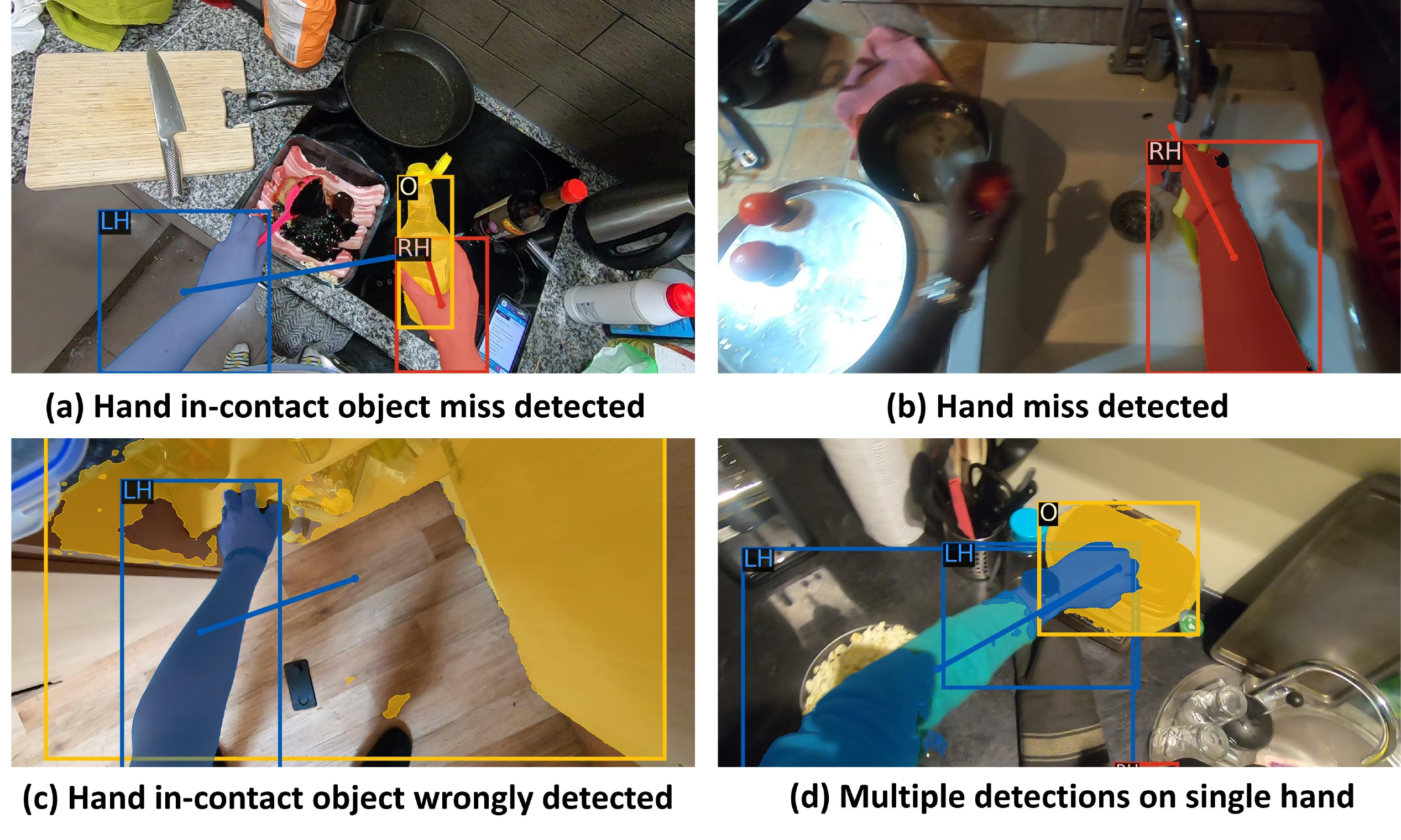}
     \caption{Illustration of the challenges in hand object segmentation on EPIC-KITCHENS VISOR dataset. (a) Missed detection of hand in-contact objects due to partial occlusion caused by objects like sauce on a spoon. (b) Hand misidentification due to blurred hand images. (c) Incorrect detection of hand in-contact objects where the majority of the object is occluded by the hand. (d) Single hand being detected as multiple parts due to strong contrast.   
     To address these various challenges, we propose an extension to the baseline model that incorporates learning for both object segmentation and detection tasks. Additionally, we introduce hand-crafted constraints designed to effectively merge the outputs of segmentation, detection, and SAM.
     }
     \label{figure:intro}
\end{figure}

VISOR \cite{VISOR2022}, a new dataset containing pixel-wise annotations, introduces a cutting-edge benchmark specifically designed for segmenting hands and active objects within the context of EPIC-KITCHENS egocentric videos \cite{damen2018scalling}. The dataset consists of various annotations including 272K manually annotated semantic masks representing 257 object classes, 9.9M interpolated dense masks, and 67K hand-object relations. It encompasses a total of 36 hours of 179 untrimmed videos. Specifically, there are 115 videos allocated for training, 43 videos (including 5 unseen ones) designated for validation, and 21 videos (including 4 unseen ones) reserved for testing purposes. The Hand Object Segmentation (HOS) challenge utilizes the dataset to tackle the task of segmenting instances of hands and objects that are in contact with each other. In addition to segmenting the hand and object regions, the challenge also requires participants to predict the hand side (left/right), the contact state (contact/no-contact), and the offset between the hand and the contacted object. The baseline method PointRend \cite{pointrend} demonstrates accurate hand segmentation, achieving an average precision (AP) of 0.909 and 0.954 on the validation and test sets, respectively. However, the task of segmenting hands in contact with objects presents significant challenges. The baseline method only achieves an AP of 0.307 and 0.337 on the validation and test sets for this specific task.  In Fig~\ref{figure:intro}, we present some challenging samples from the EPIC-KITCHENS VISOR dataset. As depicted in the figure, the detection of hands and objects may encounter challenges such as partial occlusion, motion blur, or intense variations in lighting, leading to potential misidentification or incorrect detection. 

\begin{figure}[!t]
  \centering
  \centerline{\includegraphics[width=1.0\linewidth]{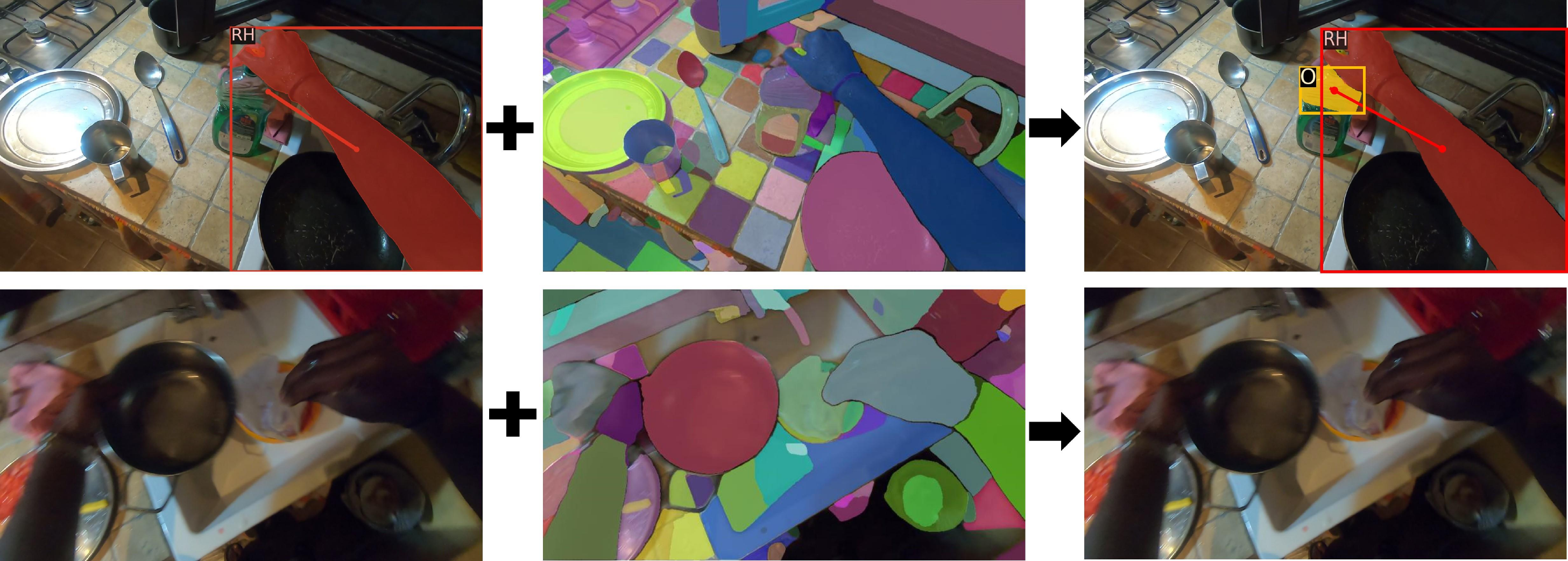}}
\caption{Samples that demonstrate whether the segmentation model can offer valuable information for extracting the desired hand or object segmentation.}
\label{fig:sam_example}
\end{figure}

Recently, a large foundation model for image segmentation is released by Facebook, achieving high performance in zero-shot segmentation \cite{SAM}. We leverage its capabilities to significantly improve segmentation accuracy. While SAM has successfully learned a broad understanding of objects and can generate masks for various objects (including unseen objects) in images or videos, it relies on prompts (\eg clicking or drawing box) to indicate the desired object mask. Meanwhile, SAM's output is class-agnostic, meaning it lacks the ability to specifically identify and segment user-specified objects without guidance.  In certain instances, the results obtained from the segmentation model can offer valuable guidance for improving segmentation derived from SAM's output. As illustrated in the first row of Fig. \ref{fig:sam_example}, despite the failure to segment the dish soap bottle accurately, the predicted contact information can be utilized to extract a partial mask of the bottle. However, in cases where the segmentation model fails to provide any useful information, as depicted in the second row of Fig. \ref{fig:sam_example}, even though SAM exhibits accurate segmentation capabilities, it struggles to extract the hands and in-contact objects. 

Object detection aims to identify and localize objects within an image, while instance segmentation goes a step further by providing precise instance masks at the pixel level, along with their corresponding class names \cite{segmentation-survey}. This additional level of detail enables instance segmentation to solve both the tasks of object detection and semantic segmentation simultaneously.  Hence, object detection is generally considered less computationally demanding than object segmentation when applied to the same training set. By leveraging SAM's ability to utilize bounding boxes as prompts, we propose a solution to tackle missed hand and object segmentation. This involves training an object detection model and implementing tailored constraints for extracting masks from the outputs of segmentation, detection and SAM models.

\begin{figure*}[t]
  \centering
  \includegraphics[width=0.9\linewidth]{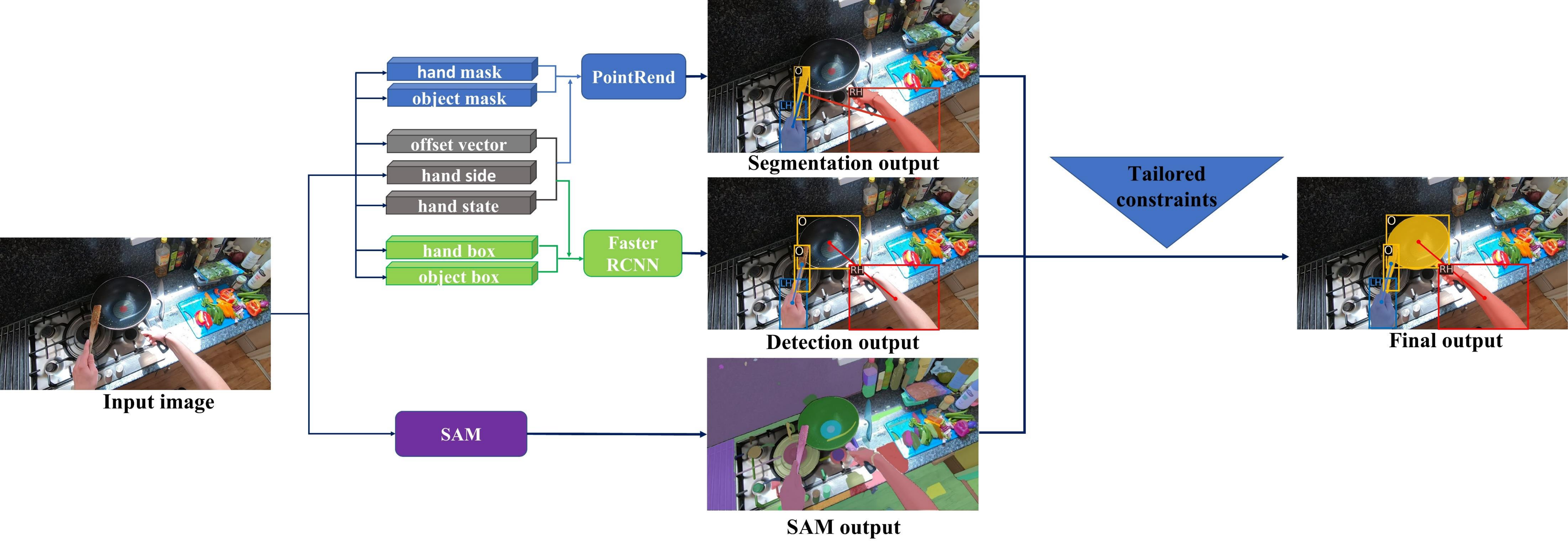}
  \caption{Overall architecture of the proposed framework. For the hand and object segmentation task, we employ the PointRend networks, utilizing hand and object mask annotations along with hand side, hand state, and offset vector. On the other hand, for the hand and object detection task, we utilize the Faster-RCNN training framework, incorporating hand and object boxes along with hand side, hand state, and offset vector. The segmentation and detection outputs are then fused with SAM's output and subjected to tailored constraints to generate the final output.
  This is best viewed in color.} 
\label{fig:architect}
\end{figure*}

The architecture of our proposed method is shown in Fig. \ref{fig:architect}. The system comprises three modules: object segmentation, object detection, and SAM module. The object segmentation module is trained using PointRend neural network, with hand and object masks as inputs. The object detection module is trained using the Faster-RCNN framework \cite{fasterrcnn}, utilizing hand and object boxes as inputs. Both modules are also trained to predict hand side (left or right), hand state (in-contact or non-contact), and offset vectors, in addition to detecting and segmenting hands and objects. The SAM module generates zero-shot object, hand, and scene segmentation results. By applying customized constraints to these three outputs, the module produces the final segmentation results for hands and objects. 

\section{Our Approach}
In this section, we will delve into the technical intricacies of our proposed approach. Our overall architecture, as depicted in Fig.~\ref{fig:architect}, consists of three main modules: PointRend segmentation, Faster-RCNN detection and SAM module, and a tailored constraints algorithm. For the SAM model, we employ it for zero-shot segmentation on the VISOR dataset without retraining it, so we do not delve into technical details of this module. 
These modules work in tandem to extract the final segmentation from the outputs they generate.

\subsection{PointRend Hand and Object Segmentation}
\label{sec:pointrend}
Following the baseline provided for the HOS task by the organizers, we utilized the PointRend instance segmentation model \cite{pointrend} implemented in Detectron2 \cite{detectron2}. The model was equipped with an R50-FPN backbone and trained using the standard 1× learning rate scheduled configuration by default. Training was performed for 90,000 iterations with a batch size of 24 and a base learning rate of 0.02.
To cater specifically to the HOS task, we incorporated three additional linear layers as heads after the ROI-Pooled feature. These heads were designed to independently predict hand side (feature size: 2), contact state (feature size: 2), and offset (feature size: 3).
During the training process, we employed Cross-Entropy loss for hand side and contact state prediction, while Mean Squared Error (MSE) loss was utilized for offset prediction.

According to \cite{VISOR2022}, the model achieved accurate segmentation of hands when tested on a separate test set, with the side being easier to predict compared to the contact state. However, it is challenging to segment the objects accurately due to the long tail distribution of object classes and heavy occlusion. 
Meanwhile, there are several challenges on predicting hand contact state and segmenting objects in contact, including the difficulty in distinguishing hand overlap and contact, the diverse range of held objects, and occlusion caused by interactions between hands and objects or between multiple hands. Typically, leveraging multiple frames extracted densely from a video can enhance performance in these tasks. However, in the case of the test set for the Hand-Object Segmentation (HOS) task, images are sparsely extracted from videos, limiting the potential benefit from utilizing the continuity of action to improve performance.
 \\

\subsection{Faster-RCNN Hand and Object Detection}
\label{sec:Faster-rcnn}
Similar to the approach in \cite{hos-Detector}, our system is constructed upon a well-established object detection framework, Faster-RCNN \cite{fasterrcnn}, with the inclusion of auxiliary predictions and losses per-bounding box. We intentionally selected Faster-RCNN as the foundation for its recognized performance in detection tasks, and any enhancements made to the base network are independent of our contributions. We utilize a ResNet-101 \cite{Resnet} backbone as the core architecture for our model. The backbone is trained for 10 epochs, employing a learning rate of 0.001, and utilizing a batch size of 4. Specifically, we extend Faster-RCNN's capabilities to detect two specific object classes: human hands and contacted objects. The bounding box annotations for both hand and object can be readily derived from their respective mask annotations. The network, akin to standard Faster-RCNN, predicts whether each anchor box contains an object, determines its category, and refines the bounding box regression adjustments—these aspects remain unchanged. Moreover, similar to the PointRend segmentation model, we incorporate auxiliary outputs (hand side, hand state and offset vector) within the Faster-RCNN model. These auxiliary outputs are derived from the same ROI-pooled features used for standard classification outputs.  

\begin{figure}[!t]
  \centering
  \centerline{\includegraphics[width=1.0\linewidth]{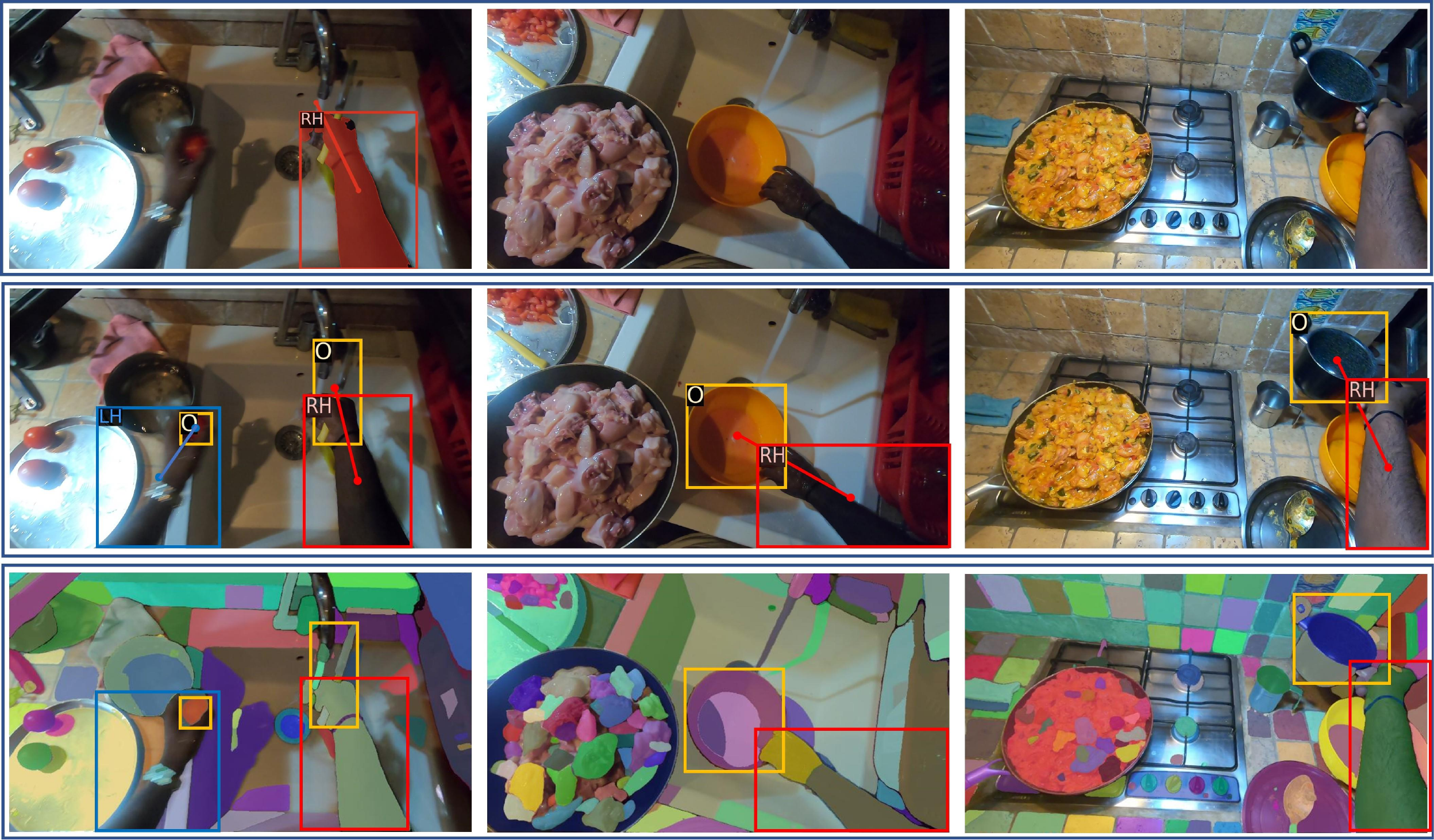}}
\caption{Performance comparison among segmentation (first row), detection (second row) and SAM model (last row). It is evident from the results that the detection model performs better than the segmentation model when trained on the same samples.}
\label{fig:detection_effectiveness}
\end{figure}

By considering the detection task as a form of dimension reduction for the segmentation task on a specific dataset, we can potentially achieve improved performance according to their respective evaluation metrics. This can be supported by visualized samples in Fig. \ref{fig:detection_effectiveness}. The first two rows showcase the segmentation and detection results on the same input images. It is evident that the detection model is capable of identifying some hand and object instances that were initially missed by the segmentation model. The third row demonstrates the impressive class-agnostic SAM output of the input images. This implies that with accurate bounding boxes for hands and objects, there is a high probability of successfully segmenting their respective masks.

\subsection{Tailored Constraints for Mask Selection}
\label{sec:tailored-constraints}
Selecting the desired object masks from a specific region of the SAM output is not a straightforward task. The difficulties  mainly arise from two aspects. Firstly, the mask of the desired object may not be the dominant region within the specified area, as object shapes can vary significantly. An illustrative example is given in the first image of Fig. \ref{fig:detection_effectiveness}, where the right hand's in-contact object (a knife) does not occupy the largest portion within the yellow box. Secondly, the SAM output itself may suffer from over-segmentation on objects. An instance of this can be seen in the first row of Fig. \ref{fig:sam_example}, where the dish soap bottle is segmented into four separate parts. Hence, we developed an algorithm that incorporates customized constraints specifically designed to address these challenges and thereby generating accurate object masks. In essence, our algorithm adheres to the following rules: (i) It prioritizes the selection of the hand mask followed by the object mask.
(ii) Pixels belonging to the hand or object, as identified by the segmentation model, are given higher priority for inclusion in the final output.  The detail algorithm is depicted in Algorithm \ref{ag:constraints}.

\begin{algorithm}
\SetAlgoLined
\textit{H} for hand, \textit{O} for object, \textit{P} for parts in SAM \;
Constraints used for mask selection\;
 C1: whether the center of the region is on \textit{P} (0 or 1)\;
 C2: the ratio of \textit{P} in the region\;
 C3: the ratio of the region pixels on \textit{P}\;
 \For{each input image}{
     Seg=segmentation\_model(input) \;
     Det=detection\_model(input) \;
     Sam=SAM\_model(input) \;
     
     \eIf{H $\in$ Det and H $\exists$ Seg}{     
        \eIf{iou(box\_H\_det, box\_H\_seg) $>$ iou\_th}
       {H\_final=H\_seg\;}       
       {H\_id=iou(H\_seg,P\_sam).index(max(iou(H\_seg,P\_sam)))\;
       H\_final=P\_sam[H\_id]\;
       }
     
     }{\eIf{H $\in$ Det and H  not $\exists$ Seg}{
     \For{each P in box\_H\_det}{F=w1*C1+w2*C2+w3*C3\;  
      H\_id=F.index(max(F))\;
      H\_final=P\_sam[H\_id]\;}
     }{continue\;}}
     \eIf{O $\in$ Det and O $\exists$ Seg}{     
        \eIf{iou(box\_O\_det, box\_O\_seg) $>$ iou\_th}
       {O\_final=O\_seg\;}       {Remove confirmed H from P\_sam\;
       O\_id=iou(O\_seg,P\_sam).index(max(iou(O\_seg,P\_sam)))\;
       O\_final=P\_sam[O\_id]\;       
       }
     
     }{\eIf{O $\in$ Det and O not $\exists$ Seg}{\eIf{contact point detected}{
       O\_final= P(contact point $\in$ P)}{F=w1*C1+w2*C2+w3*C3\;
        O\_id=F.index(max(F))\;
        O\_final=P\_sam[O\_id]\;}}{continue\;}}
     }
\caption{Mask selection in a region}
\label{ag:constraints}
\end{algorithm}

\section{Experiments}

\subsection{Datasets}
\label{sec:dataset}
The EPIC-VISOR dataset (VISOR2022) consists of a training set comprising 32,857 images distributed across 242 classes. Additionally, there is a validation set containing 7,747 images from 182 classes, with 9 of these classes being unseen during training. Lastly, the dataset includes a test set consisting of 10,125 images belonging to 160 classes, among which 6 classes are unseen during training.

\begin{table*}[ht]
\caption{The performance comparison between the baseline method and our method on the EPIC-KITCHENS-VISOR test set.}
\vspace{2mm}
\centering
\scalebox{0.9}{
\begin{tabular}{cccccc}
\hline
 & \textbf{Hand} & \textbf{Hand + Side} & \textbf{Hand + Contact} & \textbf{Object} & \textbf{Hand + Object} \\ \hline
\textbf{Baseline} & 0.9541 & 0.9238 & 0.7870 & 0.3373 & 0.60 \\ \hline
\textbf{Ours} & 0.9566 & 0.9515 & 0.7873 & 0.3511 & 0.6485 \\ \hline
\end{tabular}
}
\label{tab:results_comparision}
\end{table*}

\subsection{Implementation Details}
\label{sec:implement-detail}
The training detail of PointRend segmentation model and Faster-RCNN detection model are illustrated in sections \ref{sec:pointrend} and \ref{sec:Faster-rcnn}. For the SAM model, we used the ViT-H version. 
All models and algorithms are trained and tested on a PyTorch platform using a machine equipped with 4 NVIDIA GeForce GPUs. \\
The evaluation of hand performance encompasses three schemes: (1) Hand Segmentation: This scheme focuses on hand segmentation tasks, utilizing the COCO Mask AP \cite{COCO} as the evaluation metric. (2) Hand + Side: In addition to hand segmentation, this scheme involves hand side classification, distinguishing between left and right hands. (3) Hand + Contact: This scheme involves hand state classification, determining whether the hand is in contact or not. The performance evaluation of in-contact object segmentation task is conducted using the Mask AP metric. To further assess the explicit association between hands and in-contact objects, each hand is combined with its corresponding in-contact object mask (if applicable) and evaluate them as a single class referred to as ``Hand+Object". This allows for a comprehensive evaluation of the joint performance of hands and in-contact objects.

\subsection{Results}
\label{Sec:results}
We show the performance comparison between baseline segmentation model and our method on the test set in Table \ref{tab:results_comparision}. Our method achieves marginal improvement on the hand segmentation task. This can be attributed to the fact that the baseline method already accurately segments the hands, achieving a high AP 0.954. Consequently, the scope for further enhancement is limited. For the in-contact object segmentation task, our method yields a gain of 1.4\% in AP, which is noteworthy considering the test set included a total of 160 object classes, including 6 unseen classes. This achievement underscores the impressive nature of our AP gain. For hand side classification, our method exhibits an approximately 3\% improvement in performance. This enhancement can potentially be attributed to the compensation provided by our detection model. In terms of hand state classification, our method achieves a slight improvement in accuracy over the baseline method. On hand and object combination, our method demonstrates a significant performance improvement of approximately 4.7\%. This indicates that our approach is capable of delivering enhanced overall performance for both hand and in-contact objects in a joint manner.

\section{Conclusion}
In this report, we describe the technical details of our approach to the EPIC-KITCHENS VISOR HOS Challenge. Specifically, we propose a solution to tackle missed hand and object segmentation by involving training an object detection model and implementing tailored constraints for extracting masks from the outputs of segmentation, detection and SAM models. The experimental results on the EPIC-KITCHENS VISOR dataset demonstrate the effectiveness of our proposed method. Notably, our final submission ranks first on the leaderboard in terms of average of COCO Mask AP evaluation metric on five tasks (e.g., Hand segmentation, Hand+Side, Hand+Contact, Object segmentation, Hand+Object).

\section*{Acknowledgments}
We would like to thank Dr. Joo Hwee Lim for his support and guidance. This research is supported by the Agency for Science, Technology and Research (A*STAR) under its AME Programmatic Funding Scheme (Project \#A18A2b0046).

{\small
\bibliographystyle{ieee_fullname}
\bibliography{egbib}
}

\end{document}